\documentclass[sigconf,anonymous=False]{acmart}

\AtBeginDocument{%
  \providecommand\BibTeX{{%
    \normalfont B\kern-0.5em{\scshape i\kern-0.25em b}\kern-0.8em\TeX}}}
    
\usepackage{graphicx}
\usepackage{subcaption}
\usepackage{wrapfig}
\usepackage{dsfont}
\usepackage{textgreek}
\usepackage{scalerel}
\usepackage{subcaption}

\setcopyright{acmcopyright}
\copyrightyear{2021}
\acmYear{2021}
\acmDOI{_}

\acmConference[Preprint]{}{currently under review.}{}
\acmPrice{15.00}
\acmISBN{978-1-4503-XXXX-X/18/06}

\begin{document}

\title{Multi-view Contrastive Self-Supervised Learning of \\ Accounting Data Representations for Downstream Audit Tasks}

\author{Marco Schreyer}
\affiliation{%
    \institution{University of St. Gallen}
    \city{St. Gallen}
    \country{Switzerland}
}
\email{marco.schreyer@unisg.ch}

\author{Timur Sattarov}
\affiliation{%
    \institution{Deutsche Bundesbank}
    \city{Frankfurt am Main}
    \country{Germany}
}
\email{timur.sattarov@bundesbank.de}



\author{Damian Borth}
\affiliation{%
    \institution{University of St. Gallen}
    \city{St. Gallen}
    \country{Switzerland}
}
\email{damian.borth@unisg.ch}


\newcommand{\Tau}{\mathrm{T}}

\renewcommand{\shortauthors}{Schreyer and Sattarov, et al.}

\begin{abstract}

International audit standards require the direct assessment of a financial statement's underlying accounting transactions, referred to as journal entries. Recently, driven by the advances in artificial intelligence, deep learning inspired audit techniques have emerged in the field of auditing vast quantities of journal entry data. Nowadays, the majority of such methods rely on a set of specialized models, each trained for a particular audit task. At the same time, when conducting a financial statement audit, audit teams are confronted with (i) challenging time-budget constraints, (ii) extensive documentation obligations, and (iii) strict model interpretability requirements. As a result, auditors prefer to harness only a single preferably `multi-purpose' model throughout an audit engagement. We propose a contrastive self-supervised learning framework designed to learn audit task invariant accounting data representations to meet this requirement. The framework encompasses deliberate interacting data augmentation policies that utilize the attribute characteristics of journal entry data. We evaluate the framework on two real-world datasets of city payments and transfer the learned representations to three downstream audit tasks: anomaly detection, audit sampling, and audit documentation. Our experimental results provide empirical evidence that the proposed framework offers the ability to increase the efficiency of audits by learning rich and interpretable `multi-task' representations.

\end{abstract}

\ccsdesc[300]{Computing methodologies~Machine learning}
\ccsdesc[300]{Computing methodologies~Multi-task learning}
\ccsdesc[300]{Computing methodologies~Dimensionality reduction and manifold learning}
\ccsdesc[300]{Information systems~Enterprise resource planning}

\keywords{representation learning, self-supervised learning, multi-task learning, audit, computer assisted audit techniques, accounting information systems, enterprise resource planning systems}

\maketitle

\section{Introduction}
\label{sec:introduction}
\begin{figure}[t!]
	\hspace*{0.0cm} \includegraphics[width=8.0cm, angle=0, trim={1.0cm 0.0cm 0.0 0.0}]{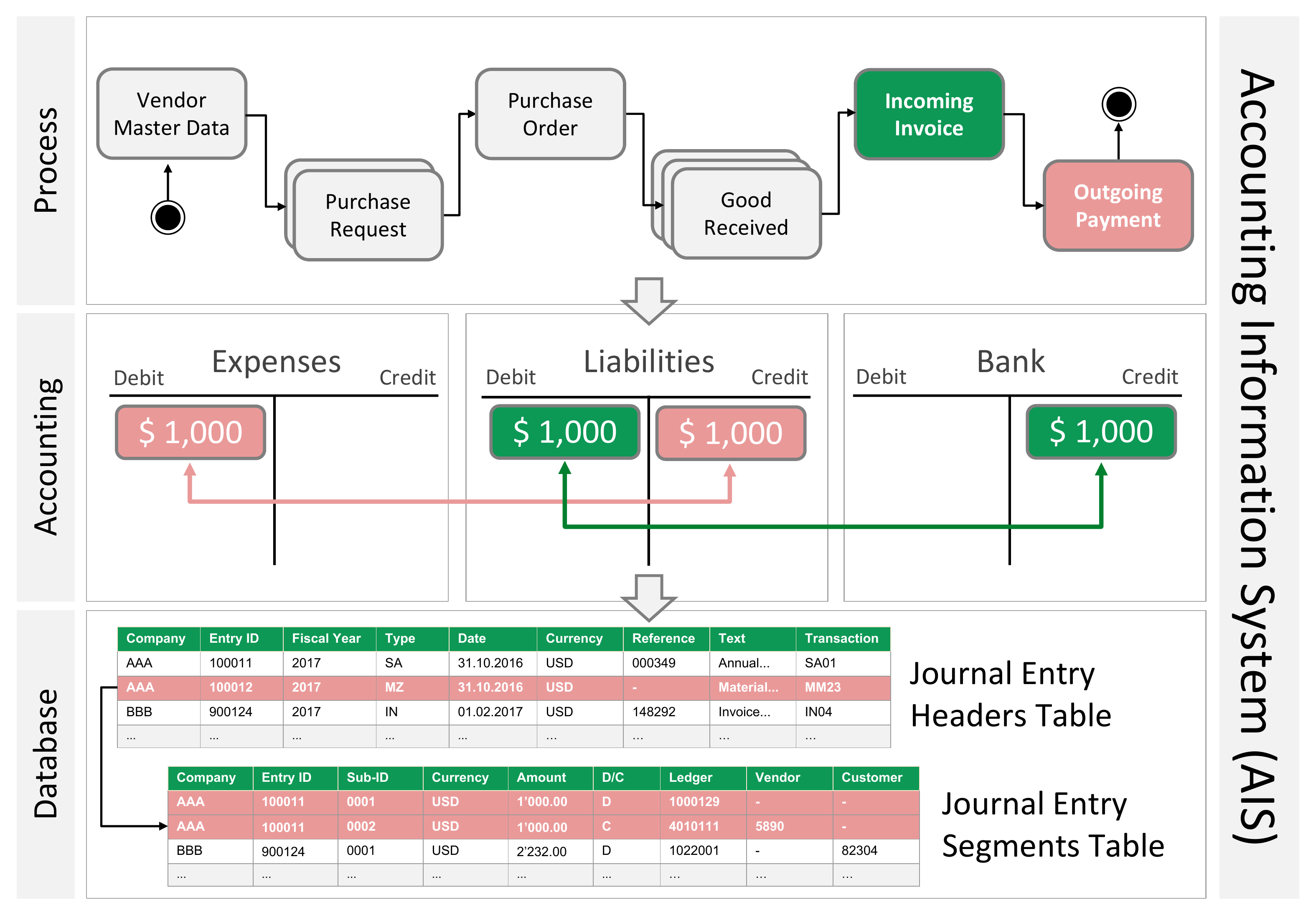}
	\vspace{-3mm}
	\caption{Hierarchical view of an Accounting Information System (AIS) that records distinct layer of abstractions, namely (1) the business process, (2) the accounting and (3) technical journal entry information in designated tables.}
	\label{fig:ais_system}
	\vspace{-3mm}
\end{figure}

In general, auditors are mandated to collect reasonable assurance that an issued financial statement is free from material misstatement (\textit{'true and fair presentation`}) \citep{sas99, ifac2009}. Traditionally, auditing has been viewed as one of the cornerstones of trustworthy financial statements. Independent financial audits seek to reduce information asymmetries between management and stakeholders. As a result, the audit opinion of external auditors plays a fundamental role in the economic decision making of investors \citep{ias2007}. To detect possible misstatements, international audit standards demand the direct assessment of a statement's underlying accounting transactions, usually referred to as \textit{'journal entries`} \citep{caq2018}. Nowadays, organizations record vast quantities of such journal entries in \textit{Accounting Information Systems (AIS)} or more general \textit{Enterprise Resource Planning (ERP)} systems \citep{grabski2011}. Figure \ref{fig:ais_system} depicts an exemplary hierarchical view of the journal entry recording process in designated database tables of an ERP system.

When auditing detailed journal entry data, auditors are constantly forced to trade off the time dedicated to the audit, and its quality \citep{bowrin2010}. In recent years accounting firms have responded with technological innovations to increase audit efficiency and mitigate the audit's time pressure \citep{kokina2017}. Driven by such efforts and the rapid technological advances of artificial intelligence, deep learning \citep{LeCun2015} techniques are applied in external and internal financial audits \citep{sun2019, nonnenmacher2021b}. Nowadays, the majority of deep learning-inspired audit techniques rely on several models, each highly specialized towards a particular audit task, e.g., the detection of anomalous journal entries \citep{schreyer2017} or to learn a representative audit sample \citep{schreyer2020}. Caused by the characteristics of modern audit engagements, such as (i) challenging time-budget constraints, (ii) extensive documentation obligations, and (iii) strict model interpretability requirements, auditors often desire to train and apply only a single (or limited number of) model(s) to audit vast quantities of accounting records. This observation motivates the design of techniques that learn invariant accounting data representations transferable across different audit tasks. Learning a single set of `general-purpose' representations potentially provides the ability to generalize better on a single task.


Learning task invariant representations without human supervision is a long-standing challenge in computer vision. Recently, the idea of \textit{contrastive self-supervised learning} \citep{oord2018, tian2020a, misra2020, chen2020, he2020} has emerged as a promising avenue of general-purpose representation learning, achieving state-of-the-art performance on the ImageNet classification challenge \cite{deng2009} and closing the gap between supervised and unsupervised training. \textit{Self-supervised learning (SSL)} \citep{doersch2015, wang2015, agrawal2015} exploits labels that can be `freely' derived from the data and uses them as intrinsic reward signals to learn transferable representations. \textit{Contrastive learning (CL)} \citep{hadsell2006} retains `rich' representations from multiple views of a single data sample by letting them attract each other while at the same time repelling them from other samples. Such views are derived from different data augmentations in order to generalize to a variety of downstream tasks \cite{goyal2019}, e.g., image classification or image segmentation.

In this work, inspired by the recent successes of contrastive self-supervised learning, we investigate whether such techniques can be utilized to learn task invariant representations of accounting records? Moreover, are the learned representations transferable to different downstream audit tasks? In summary, we present the following contributions:

\begin{itemize}

\item \textbf{Trainability} - We propose a learning framework that extracts information from intra-attribute accounting data characteristics and exploits inter-attribute correlations. 

\item \textbf{Transferability} - We demonstrate that the pre-trained representations can be transferred to different downstream audit tasks and achieve high performance. 

\item \textbf{Auditability} - We illustrate that the learned latent inter- and intra-attribute semantic disentanglement allows for a visual inspection and high interpretability by human auditors.

\end{itemize}

We believe that learning general-purpose representations of accounting data for downstream audit tasks is an essential but under-explored domain in the context of auditing. We regard this work to be an initial but promising step towards the adoption of self-supervised learning techniques in internal and external audits.

The remainder of this work is structured as follows: In Section \ref{sec:relatedwork}, we provide an overview of related work. Section \ref{sec:methodology} follows with a description of the proposed methodology to learn transferable accounting data representations from vast quantities of journal entries. The experimental setup and results are outlined in Section \ref{sec:experiments} and Section \ref{sec:results}. In Section \ref{sec:summary}, the paper concludes with a summary of the current work and highlights future research directions. A reference implementation of the proposed methodology will be made available via [\textit{url redacted due to double blind review}]. 

\begin{figure*}[ht!]
    \center
    \includegraphics[height=6.0cm,trim=0 0 0 10, clip]{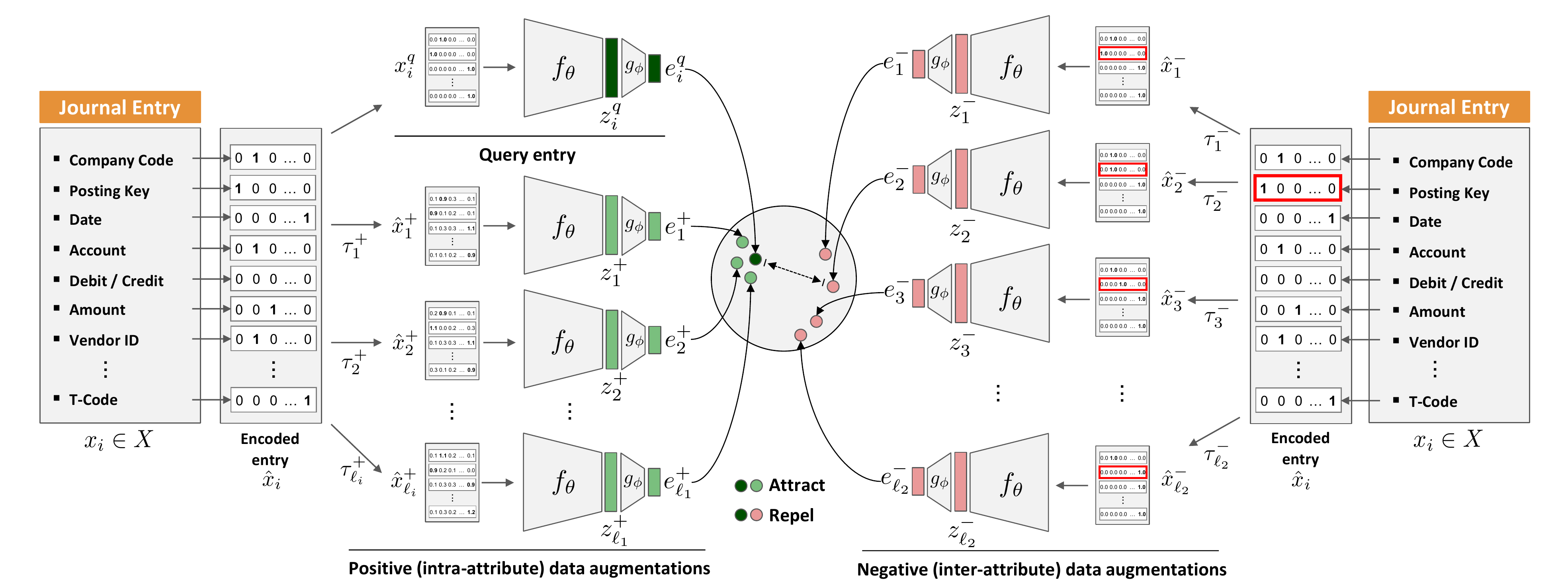}
    \vspace{-2mm}
    \caption{The adapted SimCLR contrastive learning framework \cite{chen2020}, applied to learn audit task invariant accounting journal entry representations. Given an encoded query entry $\hat{x}_{i}^{q}$, a task invariant representation $\hat{z}_{i}^{q}$ is learnt by mapping positive (intra-attribute) augmentations $\hat{x}_{i}^{+}$ of the same journal entry to nearby points in a latent space, while pushing negative (inter-attribute) augmentations $\hat{x}_{i}^{-}$ to apart points.}
    \label{fig:architecture}
    \vspace{-2mm}
\end{figure*}


\section{Related Work}
\label{sec:relatedwork}

In recent years, techniques that build on (deep) machine learning have been gradually applied to different audit tasks \citep{sun2019, cho2020}. At the same time, the idea of contrastive self-supervised learning triggered a sizable body of research by academia \citep{jing2020, jaiswal2021}. In this work, we focus our literature review on (i) learning representations of real-world accounting data and (ii) contrastive self-supervised deep learning techniques.

\subsection{Accounting Data Representation Learning}
\label{sec:accountingrepresentations}

Since the advent of AIS systems, the design of relevant data features has been of central relevance in auditing. Various techniques have been proposed that depend on human-engineered data representations, usually referred to as features, to audit structured accounting data. Such representations have been derived from: (i) specific transaction attribute characteristics \citep{Bay2002, McGlohon2009, Jans2010, thiprungsri2011, Argyrou2012, Argyrou2013}, (ii) transaction amount distributions \citep{Debreceny2010, Seow2016}, (iii) transaction or audit log information \citep{Khan2009, Khan2010, Khan2014, Islam2010}, or (iv) business process event logs \citep{Jans2011, jans2014, werner2015}. With the advent of deep learning, novel representation learning techniques have emerged in auditing. Such methods rely on `end-to-end' machine-learned accounting data representations rather than human-engineered features. Such techniques encompass: (i) autoencoder neural networks \citep{schreyer2017, schultz2020, nonnenmacher2021a}, (ii) variational autoencoders \citep{zupan2020}, and (iii) adversarial autoencoders \citep{schreyer2019a}. Lately, vector quantized variational autoencoders have been applied to learn representative audit sampling \citep{schreyer2020}. Concluding from the reviewed literature, most references either draw from human-engineered representations or learned representations highly optimized towards a particular audit task.

\subsection{Self-Supervised Representation Learning}
\label{sec:selfsupervisedlearning}

The convention of self-supervised learning (SSL) \citep{oord2018, tian2020a, misra2020, chen2020, he2020} is to solve a manually-defined `pretext' task for task invariant representation learning. In computer vision, predicting the relative location of image patches has shown to be a successful pretext task \citep{noroozi2016, dosovitskiy2015, doersch2015}. Recently, contrastive learning (CL) algorithms have been successfully employed in pretext training phases to learn invariant representations of data observations. This is implemented by minimizing a contrastive loss \citep{weinberger2009, schroff2015, oord2018} evaluated on pairs of augmented feature vectors derived from a single data observation. In computer vision, multiple image augmentation strategies have been proposed \citep{wu2018, ye2019, tian2020b, he2020, chen2020, misra2020}, e.g., rotation, cropping, random greyscale, and color jittering. Similarly, in speech analysis, several audio augmentation methods have been successfully applied \citep{park2019, wang2021, kharitonov2021}, e.g., noise augmentation, band-reject filtering, and time masking. In summary, none of the reviewed literature focuses on applying contrastive self-supervised learning in internal and external financial audits.

\vspace{1mm}

To the best of our knowledge, this work presents a first step towards learning general purpose representations of structured accounting data for downstream audit tasks.

\section{Methodology}
\label{sec:methodology}

In this section, we describe the architectural components, data augmentation techniques, and learning objective applied to learn task invariant accounting data representations. 

\subsection{Accounting Journal Entries}
\label{sec:accounting_journal_entries}

Let $X = \{x_{1}, x_{2}, ..., x_{N}\}$ be a set of $i=1, 2, ..., N$ journal entries. Each journal entry $x_{i} = \{x_{i}^{1}, x_{i}^{2}, ..., x_{i}^{M}; x_{i}^{1}, x_{i}^{2}, ..., x_{i}^{K}\}$ consists of $j=1, 2, ..., M$  categorical accounting attributes and $l=1, 2, ..., K$ numerical accounting attributes. The individual attributes describe the journal entries details, e.g., the entries' fiscal year, posting type, posting date, amount, general-ledger. Furthermore, the following pre-processing is applied to the distinct attribute classes:

\begin{itemize}

    \item The categorical attribute values \scalebox{0.9}{$x_{i}^{j}$} are converted into `one-hot' numerical tuples of bits \scalebox{0.9}{$\hat{x}_{i}^{j} \in \{0, 1\}^{\upsilon}$}, where $\upsilon$ denotes the number of unique attribute values in \scalebox{0.9}{$x^{j}$}.
    
    \item The numerical attribute values \scalebox{0.9}{$x_{i}^{l}$} are scaled, according to \scalebox{0.9}{$scale(x_{i}^{l}) = (x_{i}^{l} - \text{min}(x^{l}) / (\text{max}(x^{l}) - \text{min}(x^{l}))$}, where $\text{min}$ ($\text{max}$) denotes the minimum (maximum) attribute value in \scalebox{0.9}{$x^{l}$}.
    
\end{itemize}
 
 \noindent We denote the pre-processed journal entries $\hat{x}\in \hat{X}$, where the pre-processed attributes are defined as $\hat{x}_{i} = \{\hat{x}_{i}^{1}, \hat{x}_{i}^{2}, ..., \hat{x}_{i}^{M}; \hat{x}_{i}^{1}, \hat{x}_{i}^{2}, ..., \hat{x}_{i}^{K}\}$. 

\subsection{Contrastive Learning Framework}
\label{sec:contrastive_learnin_framework}

The objective of contrastive SSL is to learn task invariant representations by maximizing the similarity and dissimilarity over data samples which are organized in pairs. In the context of this work, pairs are created by augmenting an encoded reference journal entry $\hat{x}_{i}$, denoted as query $\hat{x}_{i}^{q}$, into: (i) one similar \textit{positive} key $\hat{x}_{i}^{+}$ and (ii) several dissimilar \textit{negative} keys $\hat{x}_{i}^{-}$. The contrastive SSL framework is designed to maximize the mutual agreement between the \textit{positive pair} $(\hat{x}_{i}^{q}, \hat{x}_{i}^{+})$ and minimize the agreement of the \textit{negative pairs} $(\hat{x}_{i}^{q}, \hat{x}_{i}^{-})$ of the same journal entry. To establish such a learning framework we utilize the recently proposed \textit{SimCLR} architecture \citep{chen2020}. The architecture, illustrated in Fig. \ref{fig:architecture}, encompasses the following components:

\begin{itemize}

    \item A stochastic \textit{data augmentation module}. The module transforms a given query entry $\hat{x}_{i}^{q}$ into $\ell_{1}$ positive ($\ell_{2}$ negative) augmentations \scalebox{0.85}{ $\mathcal{T}^{+}(\hat{x}_{i}^{q}) = \{\hat{x}_{\omega}^{+}\}^{\ell_{1}}_{\omega=1}$} (corresp. \scalebox{0.85}{$\mathcal{T}^{-}(\hat{x}_{i}^{q}) = \{\hat{x}_{\lambda}^{-}\}^{\ell_{2}}_{\lambda=1}$}) of the same entry considered as positive (negative) pairs.

    \item A neural \textit{network base encoder} $f_{\theta}(\cdot)$ with parameters $\theta$ that extracts a representation $z_{i}$ from a given encoded and augmented journal entry $\hat{x}_{i}$. Commonly a feed-forward network is used to map $\hat{x}_{i}$ to a latent space $Z \in \mathcal{R}^{d_1}$.
 
    \item A neural \textit{network projection head} $g_{\phi}(\cdot)$ with parameters $\phi$ which maps a representation $z_{i}$ to an embedding $e_{i}$ of a normalized space $H \in \mathcal{R}^{d_2}$, where $d_{2} < d_{1}$. $H$ is used to compute a contrastive objective of the learning task.
    
    \item A \textit{contrastive learning objective} that, given a query embedding $e_{i}^{q}$, aims to identify the positive key over a set of embeddings \scalebox{0.9}{ $\{ e_{k} \}_{i=1}^{\ell_{2}+1} = e_{i}^{+} \cup \{ e_{\lambda}^{-} \}^{\ell_{2}}_{\lambda=1}$} consisting of the positive embedding $e_{i}^{+}$ and a set of $\ell_{2}$ negative embeddings \scalebox{0.9}{ $\{ e_{\lambda}^{-} \}^{\ell_{2}}_{\lambda=1}$ }.
    
\end{itemize}

\subsection{Contrastive Learning Objective}
\label{sec:multiview_contrastive_learning}

A common contrastive learning objective is the \textit{InfoNCE} \cite{oord2018} loss which maps representations of positive pairs to nearby points in the embedding space $Z$ and representations of negative pairs to far apart points. To compute the loss, random mini-batches of $N$ examples are drawn, resulting in $2N$ pairwise data points per batch. Given a positive pair of embeddings $(e_{i}^{q}, e_{i}^{+})$, all other $2(N-1)$ augmented examples within the mini-batch are considered as negative examples. For each positive embedding pair of query $e_{i}^{q} = g_{\phi}(f_{\theta}(x_{i}^{q}))$ and key $e_{i, \omega}^{+} = g_{\phi}(f_{\theta}(\hat{x}_{i, \omega}^{+}))$, the InfoNCE loss is defined as: 

\begin{equation}
    \mathcal{L}^{\omega}_{\scaleto{NCE}{3pt}}(e_{i}^{q}, e_{i, \omega}^{+}, \{e_{k}\}) = - \log \frac{\exp(sim(e_{i}^{q}, e_{i, \omega}^{+}) / \tau)}{\sum^{2N}_{k=1} \mathds{1}_{[k \neq i]} \exp (sim(e_{i}^{q}, e_{k}) / \tau)},
\label{equ:single_contrastive_loss}
\end{equation} 

\noindent where $\mathds{1}$ denotes an indicator function evaluating to 1 if $k \neq i$, $\omega$ denotes an applied positive augmentation, and $\tau$ is a temperature parameter. Furthermore, $sim(u, v) = u^{T}v / \|u\|\| v\|$ denotes a normalized dot product between $u$ and $v$ (i.e. the cosine similarity). Recently, it was shown in \citep{tian2020b} that the task invariance of encoded information increases proportional to the pair-wise comparison of positive pairs. Following this observation we derive the InfoNCE loss for each query embedding $e_{i}^{q}$ over all its positive embeddings, as defined by: 

\begin{equation}
    \mathcal{L}_{\scaleto{PairNCE}{3pt}}(e_{i}^{q}, \{e_{i, \omega}^{+}\}^{\ell_{1}}_{\omega=1}, \{e_{k}\}) = \sum\nolimits_{\omega=1}^{\ell_{1}} \mathcal{L}^{\omega}_{\scaleto{NCE}{3pt}}(e_{i}^{q}, e_{i, \omega}^{+}, \{e_{k}\}),
\label{equ:multiple_contrastive_loss}
\end{equation} 

\noindent where \scalebox{0.9}{ $\{e_{\omega, i}^{+}\}^{\ell_{1}}_{\omega=1}$ } denotes the set of $\ell_{1}$ positive embeddings, $\{e_{k}\}$, and the set of all embedding keys and $\tau$ is a temperature parameter.

\subsection{Accounting Data Augmentations}
\label{sec:positive_negative_augmentations}

Learning transferable representations that result in a good downstream task performance highly relies on the applied data augmentations \citep{robinson2020}. As shown in Fig. \ref{fig:architecture}, the InfoNCE objective guides the learning framework to map positive pairs ($\hat{x}_{i}^{q}$, $\hat{x}_{i}^{+}$) to nearby locations in $Z$, and negative pairs ($\hat{x}_{i}^{q}$, $\hat{x}_{i}^{-}$) further apart. To enable the learning of such spatial relationships, we propose a data augmentation technique comprised of a positive and a negative augmentation policy, that acts directly on the unique attribute characteristics of journal entry data. 

\vspace{0.1cm}

\noindent \textbf{Negative augmentation policy:} The policy, denoted by $\mathcal{T}^{-}$, deliberately dismisses the semantic content of an entry. It is designed to create inter-attribute augmentations resulting in negative pairs ($\hat{x}_{i}^{q}$, $\hat{x}_{i}^{-}$). For a given encoded query entry $\hat{x}_{i}^{q}$, the following augmentation steps are applied to obtain a set of negative keys:

\begin{itemize} 

    \item First, the query entry $\hat{x}_{i}^{q}$ is replicated $\lambda$ times resulting in $\lambda$ negative keys \scalebox{0.9}{ $\{\hat{x}_{i}^{-}\}_{i=1}^{\lambda}$ } of the original query entry.
    
    \item Second, a random journal entry attribute $j^{*}$ is sampled from the population of categorical accounting attributes.
    
    \item Third, for each key $\hat{x}_{i}^{-}$ the high bit `1' of the attribute $j^{*}$ is swapped with another random `0` bit of the encoding.

\end{itemize}

\noindent \textbf{Positive augmentation policy:} The policy, denoted by $\mathcal{T}^{+}$, preserves the semantic content of an entry. It is designed to create intra-attribute augmentations resulting in positive pairs ($\hat{x}_{i}^{q}$, $\hat{x}_{i}^{+}$). For a given encoded query entry $\hat{x}_{i}^{q}$, the following augmentations are applied to obtain a set of positive keys:

\begin{itemize} 

    \item \textit{Random-Noise Augmentation}, in which a given encoded query entry $\hat{x}_{i}^{q}$ is perturbed by the addition of random noise \scalebox{0.95}{$\rho \sim$} \scalebox{0.95}{$\mathcal{N}(\mu_{\scaleto{R}{3pt}}, \sigma_{\scaleto{R}{3pt}})$}, formally denoted by \scalebox{0.95}{$\hat{x}_{i}^{q} + \rho$}.
    
    \item \textit{Random-Cut Augmentation}, in which the high `1' bits of a given encoded query entry $\hat{x}_{i}^{q}$ are randomly multiplied by \scalebox{0.95}{$\delta \sim \mathcal{U}(\mu_{\scaleto{C}{3pt}}, 1)$}, formally denoted by \scalebox{0.95}{$\hat{x}_{i}^{q} * \delta$}. 
    
    \item \textit{Gaussian-Kernel Augmentation}, in which a given encoded query entry $\hat{x}_{i}^{q}$ is convolved by a Gaussian kernel (`blurring'), formally denoted by \scalebox{0.95}{$\hat{x}_{i} \circledast \mathcal{N}(\mu_{\scaleto{G}{3pt}}, \sigma_{\scaleto{G}{3pt}})$}. 

\end{itemize}

\begin{figure*}[ht!]
    \center
    \includegraphics[height=6.8cm,trim=0 0 0 0, clip]{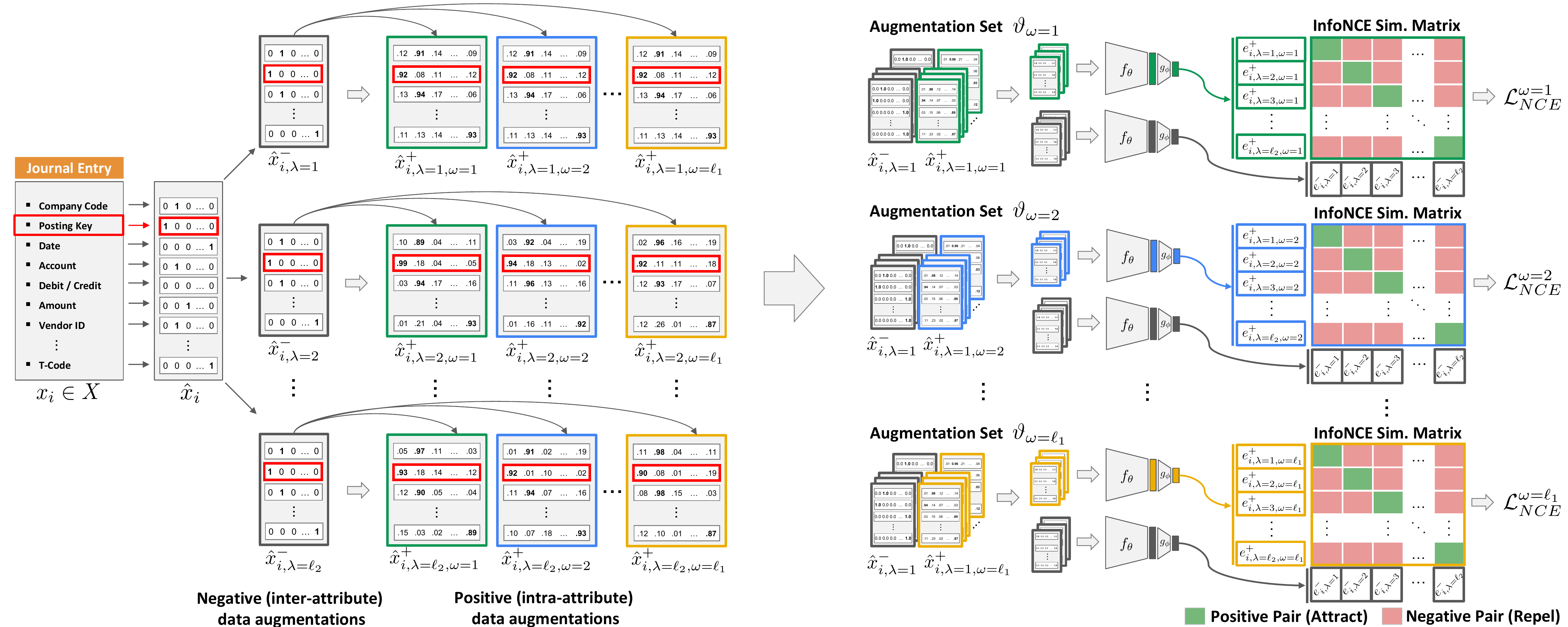}
    \vspace{-2mm}
    \caption{Overview of the proposed journal entry augmentation technique. Given an encoded journal entry $\hat{x}_{i}$, negative and positive augmentation policies are sequentially applied to yield distinct augmentation sets $\vartheta_{\omega}$. For each set, the InfoNCE loss $\mathcal{L}^{\omega}_{\scalebox{0.6}{NCE}}$ \cite{oord2018} is derived over the pair-wise embeddings. The accumulation of the set losses increases the task invariance of the learned journal entry representations.}
    \label{fig:augmentation}
    \vspace{-2mm}
\end{figure*}

\noindent Given an encoded journal entry $\hat{x}_{i}$, both augmentation polices are sequentially applied as illustrated in Fig. \ref{fig:augmentation}. First, the negative policy $\mathcal{T}^{-}$ is used to derive a set of $\ell_{2}$ inter-attribute (negative) augmentations \scalebox{0.85}{$\{\hat{x}_{i, \lambda}^{-}\}^{\ell_{2}}_{\lambda=1}$} of the entry. Second, the positive policy $\mathcal{T}^{+}$ is applied to augment each \scalebox{0.85}{$\hat{x}_{i, \lambda}^{-} \in \{ \hat{x}_{i, \lambda}^{-}\}^{\ell_{2}}_{\lambda=1}$}. Resulting in $\ell_{1}$ intra-attribute (positive) augmentations \scalebox{0.9}{$\{\hat{x}_{i, \lambda, \omega}^{+}\}^{\ell_{1}}_{\omega=1}$} for a given inter-attribute augmentation. Subsequently, for each positive augmentation $\omega$ a set of augmentation pairs is formed, formally defined by: 

\begin{equation}
    \vartheta_{\omega} = \{\hat{x}_{i, \lambda}^{-}\}^{\ell_{2}}_{\lambda=1} \cup \{\hat{x}_{i, \lambda, \omega}^{+}\}^{\ell_{2}}_{\lambda=1},
    \label{equ:augmentation_sets}
\end{equation}

\vspace{1mm}

\noindent where \scalebox{0.85}{$\{(\hat{x}_{i, \lambda_{1}}^{-}, \hat{x}_{i, \lambda_{2}, \omega}^{+}) \; | \; \lambda_{1} = \lambda_{2}\}^{\ell_{1}}_{\lambda=1}$} defines the sets' $\vartheta_{\omega}$ positive pairs, \scalebox{0.85}{$\{(\hat{x}_{i, \lambda_{1}}^{-}, \hat{x}_{i, \lambda_{2}, \omega}^{+}) \; | \; \lambda_{1} \neq \lambda_{2}\}^{\ell_{1}}_{\lambda=1}$} the sets' negative pairs, and \scalebox{0.85}{$\hat{x}_{i, \lambda_{1}}^{-}$} serves as each pair's query entry. The pair-wise InfoNCE loss, as defined by Eq. \ref{equ:multiple_contrastive_loss}, is then derived for each set of augmentation pairs $\{ \vartheta_{\omega} \}_{\omega=1}^{\ell_{1}}$. The loss forces the contrastive learning framework to map the intra-attribute (positive) pairs of each set $\vartheta_{\omega}$ to nearby locations in $Z$, and the inter-attribute (negative) pairs further apart.

\section{Experimental Setup}
\label{sec:experiments}

In this section, we describe the experimental details of the self-supervised pre-training and downstream audit task evaluation\footnote{Due to the general confidentiality of journal entry data, we evaluate the proposed methodology based on two public available real-world datasets to allow for the reproducibility of our results.}. 

\subsection{Datasets and Data Preparation}
\label{subsec:datasets}

To evaluate the invariant representation learning capabilities of the contrastive SSL framework we use two publicly available datasets of real-world financial payment data that exhibit high similarity to real-world accounting data. The datasets are referred to as \textit{dataset A} and \textit{dataset B} in the following. Dataset A corresponds to the City of Philadelphia payment data of the fiscal year 2017 \footnote{\scalebox{0.95}{\url{https://www.phila.gov/2019-03-29-philadelphias-initial-release-of-city-payments-data/}}}. It represents the city's nearly \$4.2 billion in payments obtained from almost 60 city offices, departments, boards and committees. Dataset B corresponds to vendor payments of the City of Chicago ranging from 1996 to 2020 \footnote{\scalebox{0.95}{\url{https://data.cityofchicago.org/Administration-Finance/Payments/s4vu-giwb/}}}. The data is collected from the city's 'Vendor, Contract and Payment Search' and encompasses the procurement of goods and services. The majority of attributes recorded in both datasets (similar to real-world ERP data) correspond to categorical (discrete) variables, e.g., posting date, department, vendor name, document type. We pre-process the original payment line-item attributes to (i) remove of semantically redundant attributes and (ii) obtain an encoded representation of each payment. The following descriptive statistics summarise both datasets upon successful data pre-processing: 

\begin{itemize}
\item \textbf{Dataset A:} The `City of Philadelphia' payments encompass a total of 238,894 payments comprised of $10$ categorical and one numerical attribute. The encoding resulted in a total of 8,565 encoded dimensions for each of the city's vendor payment record $\hat{x}_{i} \in \mathcal{R}^{8,565}$.

\item \textbf{Dataset B:} The `City of Chicago' payments encompass a total of 72,814 payments comprised of $7$ categorical and one numerical attribute. The encoding resulted in a total of 2,354 encoded dimensions for each of the city's vendor payment record $\hat{x}_{i} \in \mathcal{R}^{2,354}$.
\end{itemize}

\subsection{Contrastive Self-Supervised Pre-Training}
\label{subsec:training}

Our architectural setup follows the SimCLR architecture \citep{chen2020} comprised of an encoder $f_{\theta}$ and a projection-head $g_{\phi}$ as shown in Fig. \ref{fig:architecture}. The encoder uses Leaky-ReLU activation functions \cite{xu2015} with scaling factor $\alpha = 0.4$ except in the last layer where no non-linearity is applied. In the projection-head only linear transformations are applied. Table \ref{tab:architecture} depicts the architectural details of both networks. 

 \begin{table}[ht!]
  \caption{Number of neurons per layer $\eta$ of the encoder $f_{\theta}$ and projection-head $g_{\phi}$ network that comprise the \textit{SimCLR} architecture \citep{chen2020} used in our experiments.} 
  \fontsize{8}{6}\selectfont
  \centering
  \begin{tabular}{l c | c c c c c c c c c}
    \toprule
        \multicolumn{1}{l}{Net}
        & \multicolumn{1}{c}{Dataset}
        & \multicolumn{1}{c}{$\eta$ = 1}
        & \multicolumn{1}{c}{2}
        & \multicolumn{1}{c}{3}
        & \multicolumn{1}{c}{4}
        & \multicolumn{1}{c}{...}
        & \multicolumn{1}{c}{11}
        & \multicolumn{1}{c}{12}
        \\
    \midrule
    $f_{\theta}(z|\hat{x})$ & A & 4,096 & 2,048 & 1,024 & 512 & ... & 4 & 2 \\
    $g_{\phi}(e|z)$ & A & 2 & 2 & - & - & ... & - & - \\
    \midrule
    $f_{\theta}(z|\hat{x})$ & B & 2,048 & 1,024 & 512 & 256 & ... & 4 & 2 \\
    $g_{\phi}(e|z)$ & B & 2 & 2 & - & - & ... & - & - \\
    \bottomrule \\
  \end{tabular}
    \label{tab:architecture}
    \vspace*{-8mm}
 \end{table} 
 
We set $\lambda=20$ in the negative augmentation policy $\mathcal{T}^{-}$. The positive augmentation policy $\mathcal{T}^{+}$ uses the following parameters: $\mu_{\scaleto{R}{3pt}}=0.0$ and $\sigma_{\scaleto{R}{3pt}}=0.05$ (Random-Noise), $\mu_{\scaleto{C}{3pt}}=0.2$ (Random-Cut), as well as $\mu_{\scaleto{G}{3pt}}=0.0$ and $\sigma_{\scaleto{G}{3pt}}=0.8$ (Gaussian-Kernel). The parameters $\theta$ and $\phi$ of the encoder and projection-head are randomly initialized as described in \cite{glorot2010}. We train all models for a max. of 1,000 training epochs, a mini-batch size of $m=128$ journal entries, and early stopping once the loss converges. We use Adam optimization \cite{kingma2014} with $\beta_{1}=0.9$, $\beta_{2}=0.999$ and a cosine learning rate schedule in the optimization of the network parameters. Upon completion of the pre-training, we discard the projection head and transfer the learned encoder parameters $\theta^{*}$ to a given downstream audit tasks. 

\subsection{Downstream Audit Task Evaluation}
\label{subsec:downstream}

We evaluate the generalization performance of the learned journal entry representations based in three downstream audit tasks, namely (i) \textit{anomaly detection}, (ii) \textit{audit sampling}, and (iii) \textit{audit documentation}. 

\vspace*{2mm}

\noindent \textbf{Anomaly Detection:} When assessing the risk of material misstatement due to fraud, auditors are required to detect `unusual' or anomalous journal entries (ISA 240, SAS 99) \citep{isa2009b, sas99}. Thereby, two classes of anomalies can be distinguished (i) \textit{global anomalies} that exhibit unusual attribute values and (ii) \textit{local anomalies} that exhibit unusual attribute value combinations \citep{schreyer2017}. We inject a small fraction of 200 randomly sampled anomalies (60 global and 140 local) into each city payment dataset.\footnote{To create both classes of anomalies, we built upon the `Faker' project, which is publicly available via the following URL: \url{https://github.com/joke2k/faker}} To evaluate the quality of the learned representations for the purpose of anomaly detection, we utilize \textit{Autoencoder Neural Networks (AEN)} \citep{hinton2006}. The AENs encoder $f_{\theta}$ and decoder $q_{\psi}$ are of symmetrical architecture. We initialize the encoder parameters with the pre-trained $\theta^{*}$ parameters, while the decoder parameters $\psi$ are randomly initialized as described in \citep{glorot2010}. Throughout the AEN fine-tuning, the encoder parameters $\theta^{*}$ are not updated to retain the pre-trained representations. The decoder parameters are fine-tuned for a max. of 100 training epochs by minimizing a reconstruction loss, formally defined by:

\begin{equation}
    \arg \min_{\psi} = ||\hat{x}_{i} - q_{\psi}(f_{\theta^{*}}(\hat{x}_{i}))||_{2},
    \label{equ:reconstruction_loss}
\end{equation}

\noindent where $\hat{x}_{i} \in \hat{X}$ denotes an encoded payment. We use a combined loss \citep{schreyer2019a} that defines the reconstruction error of a given encoded journal entry $\hat{x}_{i}$, defined as: 

\begin{equation}
    \mathcal{L}_{\psi}( \tilde{x}_{i},\hat{x}_{i}) = \nu \hspace{1mm} \mathcal{L}^{BCE}_{\psi, cat} + (1 - \nu) \hspace{1mm} \mathcal{L}^{MSE}_{\psi, num} \;,
    \vspace{2mm}
    \label{equ:reconstruction_loss_details}
\end{equation}

\noindent where \scalebox{0.9}{$\tilde{x}_{i} = q_{\psi}(f_{\theta^{*}}(\hat{x}_{i}))$} denotes the i-\textit{th} journal entry reconstruction, \scalebox{0.8}{ $\mathcal{L}^{BCE}_{\psi, cat} = \frac{1}{N} \sum_{i=1}^{N} \tilde{x}_{i}^{cat} \cdot \log(\hat{x}_{i}^{cat}) + (1 - \tilde{x}_{i}^{cat}) \cdot \log(1 - \hat{x}_{i}^{cat})$} defines the binary-cross-entropy error over the categorical attributes, while \scalebox{0.8}{ $\mathcal{L}^{MSE}_{\psi, num} = \frac{1}{N} \sum_{i=1}^{N}{(\tilde{x}_{i}^{num} - \hat{x}_{i}^{num})}^2$} denotes the mean-squared error over the numerical attributes. To account for the higher number of categorical attributes in both datasets we set \scalebox{0.85}{$\nu=\frac{2}{3}$}. To quantitatively assess the anomaly detection capability of the fine-tuned models we derive the average precision $AP$ over the sorted payments reconstruction errors $\mathcal{L}_{\psi}$. The $AP$ summarizes the model's precision-recall curve as weighted mean of precisions at each reconstruction error threshold, formally defined by:  

\begin{equation}
    AP(\mathcal{L}_{\psi}) = \sum\nolimits_{i=1}^{N} (R_{i}- R_{i-1}) P_{i},
    \label{equ:average_precision}
\end{equation}

\noindent where $P_{i}(\mathcal{L}_{\psi})=TP/(TP + FP)$ denotes the detection precision, and $R_{i}(\mathcal{L}_{\psi})=TP /(TP + FN)$ denotes the detection recall of the i-\textit{th} reconstruction error threshold.

\vspace*{2mm}

\noindent \textbf{Audit Sampling:} When performing substantive audit procedures, auditors are required to audit a representative sample of an organization's accounting records to derive a `reasonable basis for an opinion' and mitigate sampling risks (ISA 530, SAS 39) \citep{isa2009a, sas39}. To evaluate the quality of the learned representations for the purpose of audit sampling we utilize \textit{Vector Quantized Variational Autoenoders (VQ-VAE)} \citep{vanDenOord2017}. The VQ-VAE defines a discrete latent space $Z'$ and a set of $K$ latent quantization vectors $z'_{j} \in Z'$, where $j= 1, 2, ..., K$. For a given journal entry $\hat{x}_{i}$ the VQ-VAE's encoder network $f_{\theta}$ produces a latent representation $z_{i}$. To obtain a quantization $\tilde{z}_{i}$ of $z_{i}$, a nearest neighbour lookup is performed $\tilde{z}_{i} = z'_{k}$, where $k = \arg \min_{j} ||z_{i} - z'_{j}||_{2}$. Afterwards, the quantised representation $\tilde{z}_{i}$ is passed to the decoder $q_{\psi}$. Both, encoder $f_{\theta}$ and decoder $q_{\psi}$, are of symmetrical architecture. We initialize the encoder parameters with the pre-trained $\theta^{*}$ parameters, while the decoder parameters $\psi$ are randomly initialized as described in \citep{glorot2010}. The decoder parameters are fine-tuned for a max. of 100 training epochs by minimizing a reconstruction loss, formally defined by:

\vspace*{-1mm}

\begin{equation}
\begin{aligned}
    \arg \min_{\psi} = {} & ||\hat{x}_{i} - q_{\psi}(\tilde{z}_{i})||_{2}\;+\; \alpha ||sg[z_{i}] - z'||_{2}^{2} \\[-3mm]
    &+\; \beta ||z_{i} - sg[z']||_{2}^{2}\;+\; \gamma ||\hat{x}_{i} - q_{\psi}(z_{i})||_{2}, 
\end{aligned}
\label{equ:quantization_loss}
\end{equation} 

\noindent where $z_{i}$ denotes the encoder output, $\tilde{z}_{i}$ the quantised encoder output, $z'$ the set of quantization vectors, and $sg[\cdot]$ the stop gradient operator. Throughout the VQ-VAE fine-tuning, the encoder parameters $\theta^{*}$ are not updated to retain the pre-trained representations. We use the combined reconstruction loss \citep{schreyer2019a} defined in Eq. \ref{equ:reconstruction_loss_details} to optimize the first and fourth optimization term of Eq. \ref{equ:quantization_loss}. We quantitatively assess the model's fine-tuned audit sampling capability by measuring (i) quantization perplexity and (ii) quantization purity. The average quantization perplexity defines the likelihood of a quantised representation $\tilde{z}_{i}$ of being assigned to a particular quantization $z'_{j}$, formally expressed by:

\vspace*{-1mm}

\begin{equation}
    \mathcal{P}_{erp}(\pi) = \frac{1}{K} \sum\nolimits_{j=1}^{K} 2^{- \sum_{j=1}^{K} p(\pi_{j}) \log_{2} p(\pi_{j})},
    \label{equ:perplexity_loss}
\end{equation}

\vspace*{1mm}

\noindent where \scalebox{0.9}{$\pi_{j} = \sum_{i=1}^{N} \mathds{1} [\tilde{z}_{i} = z'_{j}]$} denotes the number of payments quantized by a particular quantization and $\mathds{1}$ the indicator function. We determine the average purity of quantizations, as defined by: 

\vspace*{-1mm}

\begin{equation}
    \mathcal{P}_{urity}(\pi) = \frac{1}{K} \sum\nolimits_{j=1}^{K} 1 - \frac{|\pi_{j}'|}{\pi_{j}},
    \label{equ:purity_loss}
\end{equation}

\vspace*{1mm}

\noindent where denotes $|\pi_{j}|$ the cardinality of de-duplicated payments $\pi_{j}'$ given a particular quantization $\pi_{j}$. The payment de-duplication is conducted over the categorical attributes of each $x_{i} \in \pi_{j}$. In addition, we obtain the weighted purity \scalebox{0.9}{$\mathcal{P}'urity$} by normalizing the average purity of each quantization by the fraction of quantized payments.  

\vspace*{2mm}

\noindent \textbf{Audit Documentation:} Auditors are obliged to prepare audit documentation that enables an experienced external auditor to `understand the results of the audit procedures performed and the audit evidence obtained' (ISA 230, SAS 103) \citep{ifac2006, sas103}. As shown in Tab. \ref{tab:architecture} we restrict the encoder $f_{\theta}$ bottleneck to two dimensions for all trained models. The limitation of $Z \in \mathcal{R}^{2}$ allows for a human (visual) interpretation of the learned representations and improved documentation of associated audit findings.

\section{Experimental Results}
\label{sec:results}

In this section, we quantitatively and qualitatively assess the transferablity of the learned representations towards the downstream audit tasks. 

\begin{figure*}[ht!]
    \begin{subfigure}{0.25\textwidth}
        \centering
        \includegraphics[height=5.5cm,trim=10 10 10 28, clip]{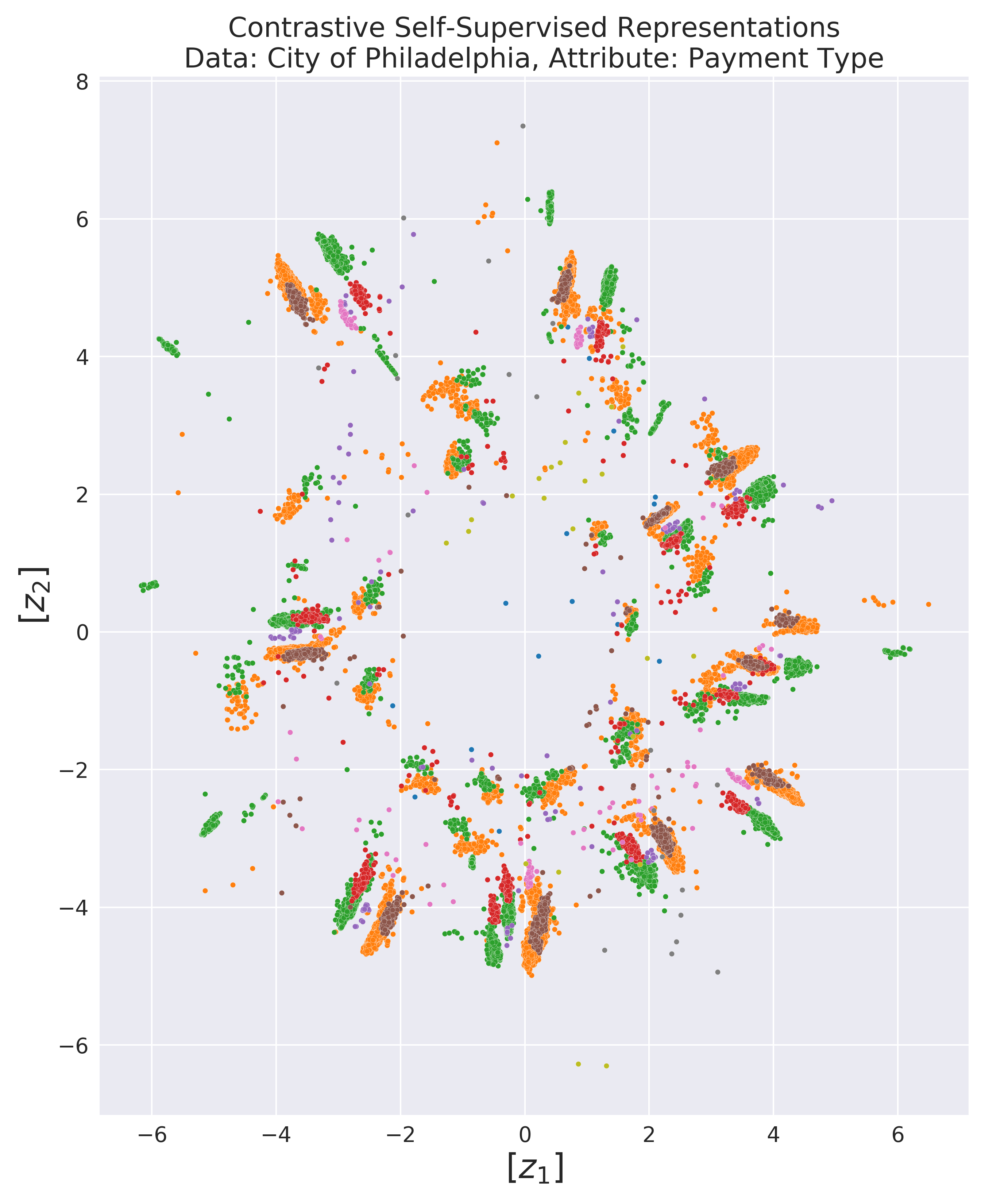}
        \vspace{-4mm}
        \caption{Attribute: Payment Type}
    \end{subfigure}
    \begin{subfigure}{0.245\textwidth}
        \centering
        \includegraphics[height=5.5cm,trim=33 10 10 28, clip]{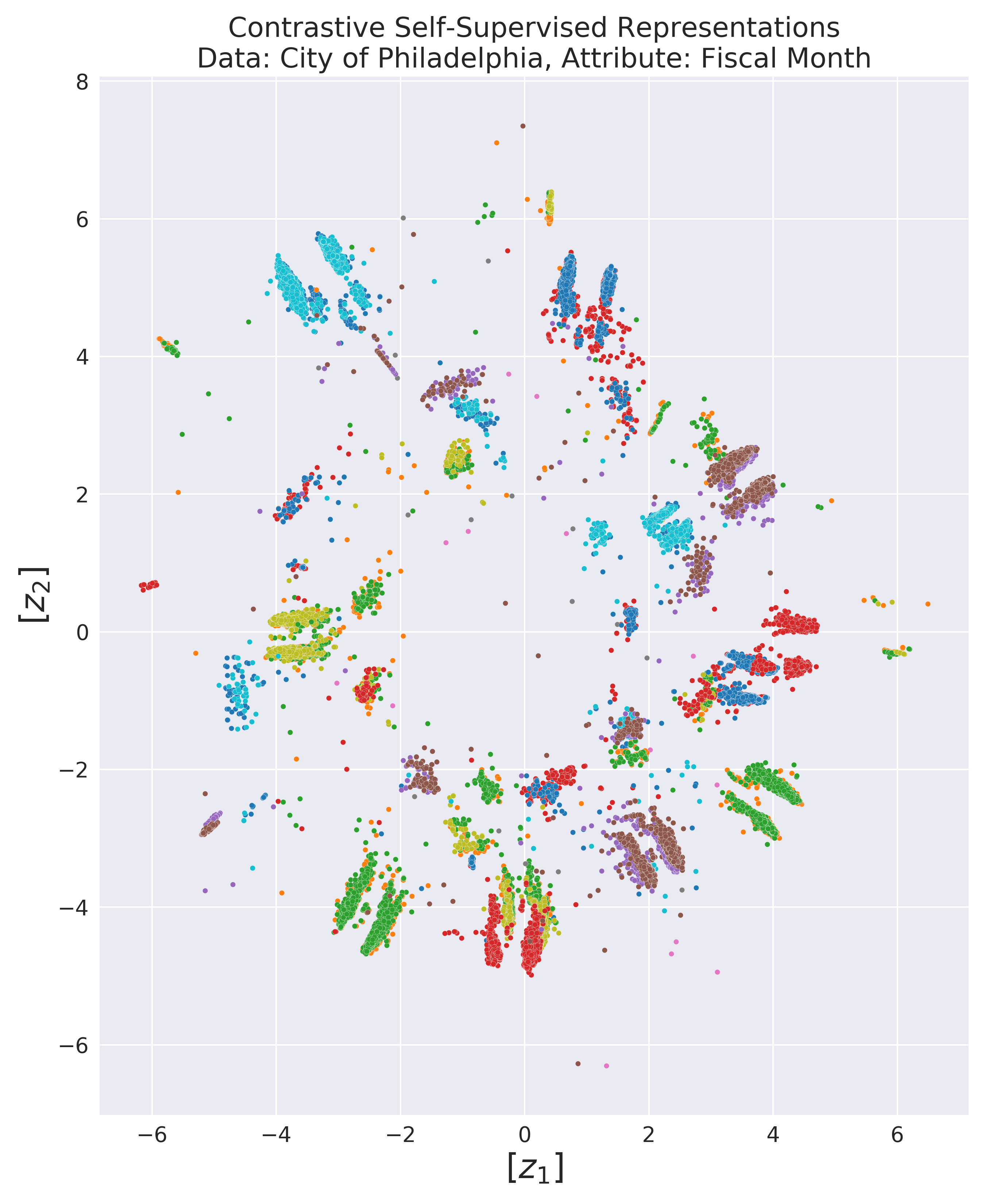}
        \vspace{-4mm}
        \caption{Attribute: Fiscal Month}
    \end{subfigure}
    \begin{subfigure}{0.245\textwidth}
        \centering
        \includegraphics[height=5.5cm,trim=33 10 10 28, clip]{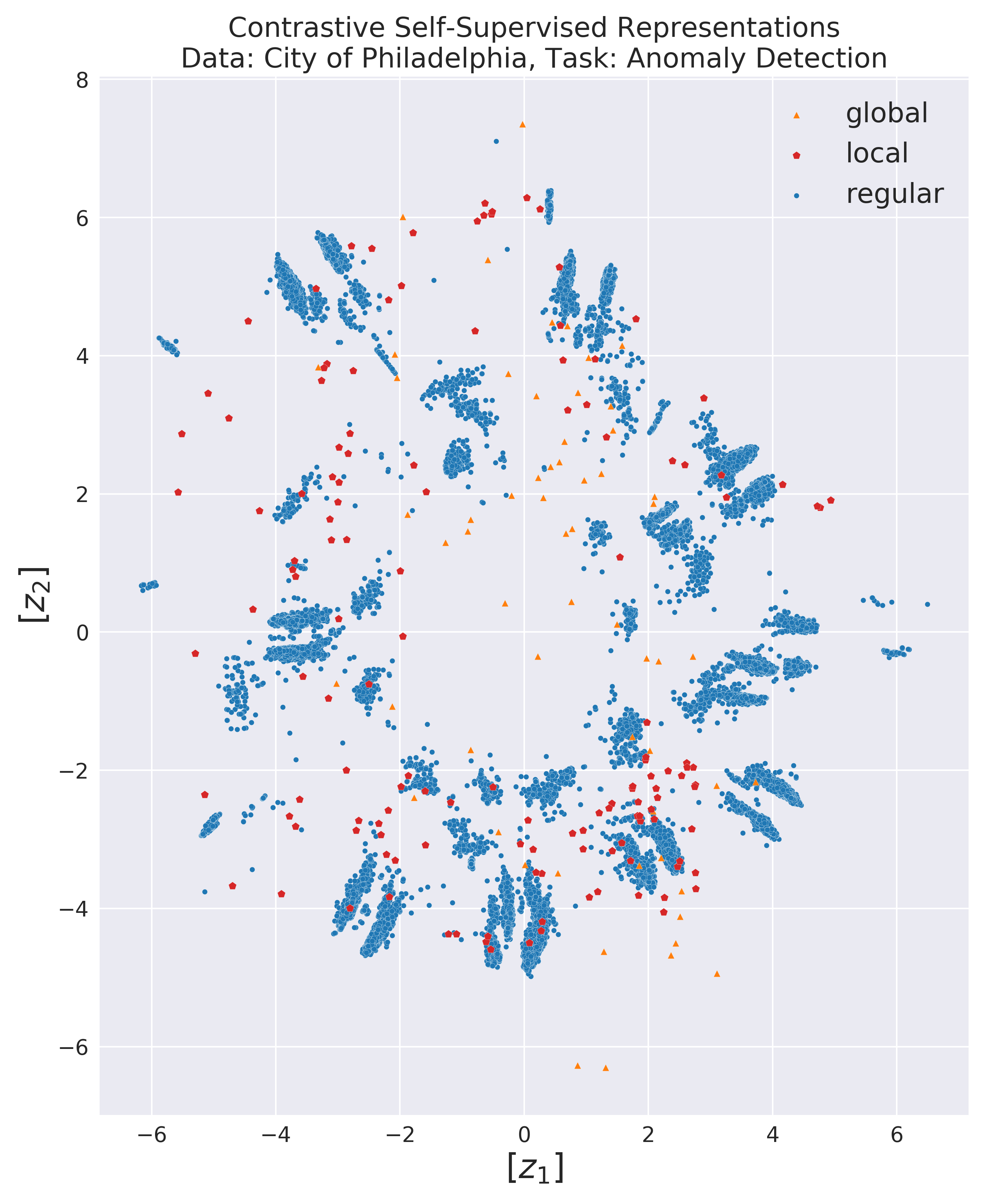}
        \vspace{-4mm}
        \caption{Anomaly Detection}
    \end{subfigure}
    \begin{subfigure}{0.245\textwidth}
        \centering
        \includegraphics[height=5.5cm,trim=33 10 10 28, clip]{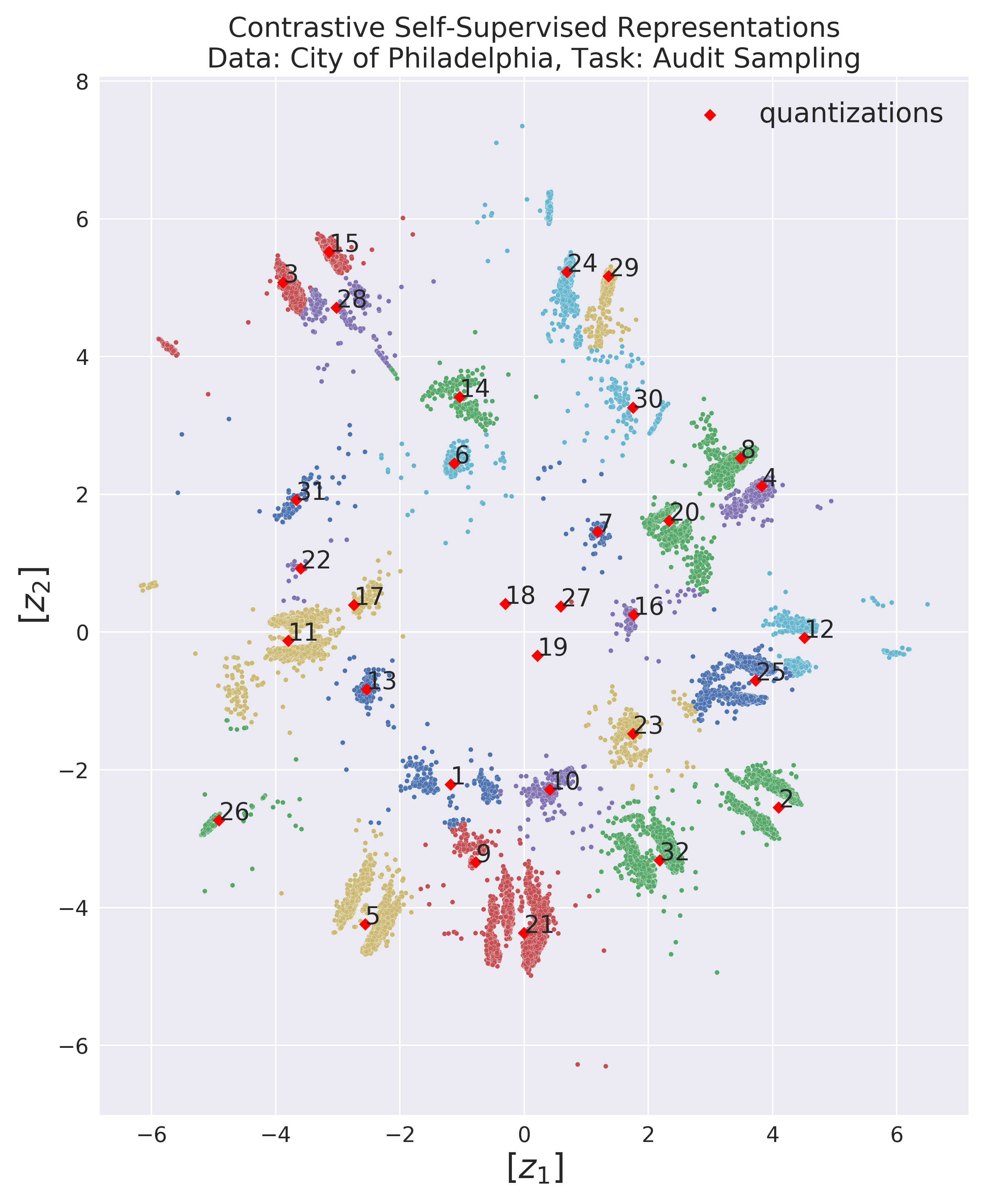}
        \vspace{-4mm}
        \caption{Audit Sampling}
    \end{subfigure}
    \vspace{-2mm}
    \caption{Learned task invariant accounting data representations $z_{i} \in \mathcal{R}^{2}$ with $\tau=0.5$ of the 238,894 `City of Philadelphia' vendor payments (dataset A). The visualisations on the left show the representations coloured according to selected payment characteristics: payment type (a) and posting month (b). The visualisations on the right show the same representations coloured according to the downstream audit task: anomaly detection (c) and audit sampling (d).}
    \label{fig:learned_representations}
    \vspace{-2mm}
\end{figure*}


\vspace*{2mm}

\noindent \textbf{Anomaly Detection:} We evaluate the fine-tuned autoencoder model (AEN\scalebox{0.7}{\textit{SSL}}) and compare its anomaly detection capability to different autoencoders (AEN) proposed in \citep{schreyer2017} which we use as a baseline. The baseline models are specifically designed for the purpose of journal entry anomaly detection. To conduct a fair comparison, we trained distinct baseline model comprised of $\eta \in [1, 6, 12]$ encoder and decoder layers for 1,000 training epochs following the training objective defined in Eq. \ref{equ:reconstruction_loss_details}.

\begin{table}[ht!]
\caption{Average precision scores obtained for all journal entry anomalies $AP_{all}$, global anomalies $AP_{global}$, local anomalies $AP_{local}$ and distinct temperature parameters $\tau$ on both city payment datasets.} 
\fontsize{8}{6}\selectfont
\centering
\begin{tabular}{l c c | r | r | r }
\toprule
    \multicolumn{1}{l}{Model}
    & \multicolumn{1}{l}{Data}
    & \multicolumn{1}{c}{$\tau$}
    & \multicolumn{1}{c}{$AP_{all}$}
    & \multicolumn{1}{c}{$AP_{global}$}
    & \multicolumn{1}{c}{$AP_{local}$}
    \\
\midrule
AEN\scalebox{0.8}{1} & A & - & 0.452 $\pm$ 0.05 & \textbf{0.404} $\pm$ \textbf{0.01} & 0.186 $\pm$ 0.01\\
AEN\scalebox{0.8}{6} & A & - & \textbf{0.587} $\pm$ \textbf{0.03} & 0.211 $\pm$ 0.08 & 0.439 $\pm$ 0.13\\
AEN\scalebox{0.8}{12} & A & - & 0.565 $\pm$ 0.02 & 0.082 $\pm$ 0.02 & \textbf{0.591} $\pm$ \textbf{0.14}\\
\midrule
AEN\scalebox{0.7}{\textit{SSL}} & A & 0.1 & 0.586 $\pm$ 0.04 & 0.371 $\pm$ 0.05 & 0.312 $\pm$ 0.03\\
AEN\scalebox{0.7}{\textit{SSL}} & A & 0.5 & 0.680 $\pm$ 0.02 & 0.412 $\pm$ 0.03 & 0.383 $\pm$ 0.02\\
AEN\scalebox{0.7}{\textit{SSL}} & A & 0.8 & \textbf{0.882} $\pm$ \textbf{0.02} & \textbf{0.673} $\pm$ \textbf{0.06} & \textbf{0.483} $\pm$ \textbf{0.03}\\
\midrule
\midrule
AEN\scalebox{0.8}{1} & B & - & 0.430 $\pm$ 0.04 & 0.051 $\pm$ 0.01 & \textbf{0.480} $\pm$ \textbf{0.04}\\
AEN\scalebox{0.8}{6} & B & - & 0.435 $\pm$ 0.10 & 0.098 $\pm$ 0.01 & 0.394 $\pm$ 0.48\\
AEN\scalebox{0.8}{12} & B & - & \textbf{0.4993} $\pm$ \textbf{0.06} & \textbf{0.113} $\pm$ \textbf{0.02} & 0.459 $\pm$ 0.08\\
\midrule
AEN\scalebox{0.7}{\textit{SSL}} & B & 0.1 & \textbf{0.472} $\pm$ \textbf{0.10} & 0.087 $\pm$ 0.01 & \textbf{0.426} $\pm$ \textbf{0.16}\\
AEN\scalebox{0.7}{\textit{SSL}} & B & 0.5 & 0.334 $\pm$ 0.05 & 0.089 $\pm$ 0.11 & 0.294 $\pm$ 0.08\\
AEN\scalebox{0.7}{\textit{SSL}} & B & 0.8 & 0.416 $\pm$ 0.13 & \textbf{0.256} $\pm$ \textbf{0.08} & 0.318 $\pm$ 0.17\\
\bottomrule \\
\multicolumn{6}{l}{\scalebox{0.8}{Variances originate from parameter initialization using five distinct random seeds.}}
\end{tabular}
\label{tab:anomaly_scores}
\end{table}

Table \ref{tab:anomaly_scores} shows the average precision scores obtained for both anomaly classes $AP_{all}$, the global anomalies $AP_{global}$, and the local anomalies $AP_{local}$ over distinct temperature parameters $\tau$. Comparing the obtained results shows that the fine-tuned contrastive SSL models yield a high overall detection precision for the distinct anomaly classes. Furthermore, the task-invariant representations demonstrate the ability to significantly outperform the baselines in both city payment datasets for the global anomalies. For the local anomalies, the task-invariant representations underperform the baselines of highly specialized AENs. In summary, the obtained results provide initial evidence of the generalization capabilities of contrastive SSL, closing the gap to learning highly specialized accounting data representations.


\vspace*{2mm}

\noindent \textbf{Audit Sampling:} We evaluate the fine-tuned VQ-VAE models (VAE\scalebox{0.7}{\textit{SSL}}) and compare its audit sampling capability to different vector-quantized VAE models (VAE) proposed in \citep{schreyer2020} which we use as a baseline. The baseline models are specifically designed for the purpose audit sampling. To conduct a fair comparison, we trained distinct baseline models comprised of different audit sample sizes $K \in [2^{3}, 2^{5}, 2^{6}]$ respectively for 1,000 training epochs following the training objective defined in Eq. \ref{equ:quantization_loss}.

\begin{table}[ht!]
\caption{Quantization purity \scalebox{0.9}{$\mathcal{P}urity$}, weighted quantization purity \scalebox{0.9}{$\mathcal{P}'urity$}, and codebook perplexity \scalebox{0.9}{$\mathcal{P}erp$} obtained for different codebook size $K$ on both city payment datasets.} 
\fontsize{8}{6}\selectfont
\centering
\begin{tabular}{c c c | r | r | r }
\toprule
    \multicolumn{1}{c}{Model}
    & \multicolumn{1}{c}{Data}
    & \multicolumn{1}{c}{$K$}
    & \multicolumn{1}{c}{\scalebox{0.9}{$\mathcal{P}urity$}}
    & \multicolumn{1}{c}{\scalebox{0.9}{$\mathcal{P}'urity$}}
    & \multicolumn{1}{c}{\scalebox{0.9}{$\mathcal{P}erp$}}
    \\
\midrule
VAE & A & $8$ & 0.174 $\pm$ 0.02 & 0.897 $\pm$ 0.01 & 6.856 $\pm$ 0.52\\
VAE & A & $64$ & 0.195 $\pm$ 0.03 & 0.974 $\pm$ 0.01 & 37.618 $\pm$ 3.88\\
VAE & A & $128$ & \textbf{0.226} $\pm$ \textbf{0.05} & \textbf{0.988} $\pm$ \textbf{0.02} & \textbf{41.251} $\pm$ \textbf{4.32}\\
\midrule
VAE\scalebox{0.7}{\textit{SSL}} & A & $8$ & 0.174 $\pm$ 0.01 & 0.897 $\pm$ 0.01 & 7.369 $\pm$ 0.45\\ 
VAE\scalebox{0.7}{\textit{SSL}} & A & $64$ & 0.313 $\pm$ 0.02 & 0.985 $\pm$ 0.01 & 26.211 $\pm$ 1.04\\
VAE\scalebox{0.7}{\textit{SSL}} & A & $128$ & \textbf{0.372} $\pm$ \textbf{0.02} & \textbf{0.992} $\pm$ \textbf{0.01} & \textbf{37.931} $\pm$ \textbf{2.21}\\
\midrule
\midrule
VAE & B & $8$ & 0.672 $\pm$ 0.01 & 0.956 $\pm$ 0.01 & 5.950 $\pm$ 0.63\\
VAE & B & $64$ & 0.756 $\pm$ 0.08 & 0.989 $\pm$ 0.01 & 21.636 $\pm$ 9.89\\
VAE & B & $128$ & \textbf{0.914} $\pm$ \textbf{0.04} & \textbf{0.997} $\pm$ \textbf{0.01} & \textbf{17.965} $\pm$ \textbf{8.01}\\
\midrule
VAE\scalebox{0.7}{\textit{SSL}} & B & $8$ & 0.955 $\pm$ 0.01 & 0.994 $\pm$ 0.01 & 5.940 $\pm$ 0.19\\
VAE\scalebox{0.7}{\textit{SSL}} & B & $64$ & 0.995 $\pm$ 0.01 & 0.999 $\pm$ 0.01 & 12.983 $\pm$ 3.47 \\
VAE\scalebox{0.7}{\textit{SSL}} & B & $128$ & \textbf{0.998} $\pm$ \textbf{0.01} & \textbf{0.999} $\pm$ \textbf{0.01} & \textbf{12.904} $\pm$ \textbf{0.25}\\
\bottomrule \\
\multicolumn{6}{l}{\scalebox{0.8}{Variances originate from parameter initialization using five distinct random seeds.}}
\end{tabular}
\label{tab:sampling_scores}
\vspace*{-3mm}
\end{table}

Table \ref{tab:sampling_scores} shows the average codebook perplexity \scalebox{0.9}{$\mathcal{P}erp$}, quantization purity \scalebox{0.9}{$\mathcal{P}urity$}, and weighted quantization purity \scalebox{0.9}{$\mathcal{P}'urity$} obtained for distinct audit sample sizes $K$. The obtained results show that the average codebook usage and quantization purity increases with increased codebook sizes. Hence, a more fine-grained quantization of the latent generative factors is learned. For both city payment datasets, the fine-tuned contrastive SSL models achieve significantly higher purity scores when compared to the baselines of specialized VQ-VAEs. This observation confirms the generalization capabilities of the pre-trained representations to yield a representative audit sample.


\vspace*{2mm}

\noindent \textbf{Audit Documentation:} We qualitatively assess the accounting-specific semantics captured by the task invariant accounting representations in different regions of the latent space $Z \in \mathcal{R}^{2}$. Figure \ref{fig:learned_representations} shows the learned representations of the City of Philadelphia Payments (Dataset A). The visualizations illustrate that the learned representations correspond to latent manifolds that `disentangle' the underlying data semantics. In the context of audit documentation the characteristics and underlying semantics of the latent space could serve as reference point for explaining deep learning inspired audit procedures and corresponding findings.


\section{Summary}
\label{sec:summary}

In this work, we proposed a self-supervised learning framework to learn task invariant accounting data representations. We demonstrated that the SimCLR architecture, enhanced by deliberately designed intra-attribute and inter-attribute data augmentation policies, yield excellent performance on the evaluated downstream audit tasks, namely (i) anomaly detection, (ii) audit sampling, and (iii) audit documentation. In summary, the obtained experimental results based on real-world datasets of city payments provide initial evidence that contrastive SSL could provide a starting point for a variety of further downstream audit procedures. Given the tremendous amount of journal entries recorded by organisations nowadays, such techniques offer the ability to increase the efficiency of audits by pre-training a single multi-purpose model and therefore mitigate the time pressure faced by auditors nowadays.

\vspace*{-2mm}


\bibliographystyle{abbrv}
\bibliography{library}

\end{document}